\documentclass[letterpaper, 10 pt, conference]{ieeeconf}  

\IEEEoverridecommandlockouts                              

\usepackage{graphics} 
\usepackage{epsfig} 
\usepackage{amsmath} 
\usepackage{amssymb}  
\usepackage{cite}
\usepackage{color}
\usepackage{balance}
\usepackage{booktabs}

\usepackage{algorithm}
\usepackage{algorithmic}

\usepackage{multirow}
\usepackage{adjustbox}

\makeatletter
\let\MYcaption\@makecaption
\makeatother

\usepackage[font=footnotesize]{subcaption}

\makeatletter
\let\@makecaption\MYcaption
\makeatother

\usepackage{subcaption}

\def\eg{{\it e.g.}}

\def\ie{{\it i.e.}}

\title{\LARGE \bf
Risk-aware Path Planning via Probabilistic Fusion of Traversability Prediction for Planetary Rovers on Heterogeneous Terrains}

\author{Masafumi Endo$^{1}$, Tatsunori Taniai$^{2}$, Ryo Yonetani$^{2}$, and Genya Ishigami$^{1}$
\thanks{*This work was partially supported by JSPS KAKENHI Grant Number JP22J22731.}
\thanks{$^{1}$Masafumi Endo and Genya Ishigami are with the Space Robotics Group, Department of Mechanical Engineering, Keio University, Kanagawa 223-8522, Japan
        {\tt\small masafumi.endo@keio.jp, 
        ishigami@mech.keio.ac.jp}
        }
\thanks{$^{2}$Tatsunori Taniai and Ryo Yonetani are with OMRON SINIC X Corporation, Tokyo 113-0033, Japan 
        \{{\tt\small tatsunori.taniai, ryo.yonetani\}@sinicx.com}
        }%
}

\begin{document}

\twocolumn[
\noindent
© 2023 IEEE. Personal use of this material is permitted. Permission from IEEE must be obtained for all other uses, in any current or future media, including reprinting/republishing this material for advertising or promotional purposes, creating new collective works, for resale or redistribution to servers or lists, or reuse of any copyrighted component of this work in other works.\\

\noindent
\textbf{Accepted article:}\\
M. Endo, T. Taniai, R. Yonetani, and G. Ishigami ``Risk-aware path planning via probabilistic fusion of traversability prediction for planetary rovers on heterogeneous terrains,'' \textit{IEEE International Conference on Robotics and Automation}, 2023.
]
\thispagestyle{empty}
\pagenumbering{gobble}
\clearpage

\maketitle
\thispagestyle{empty}
\pagestyle{empty}

\begin{abstract}
Machine learning (ML) plays a crucial role in assessing traversability for autonomous rover operations on deformable terrains but suffers from inevitable prediction errors. Especially for heterogeneous terrains where the geological features vary from place to place, erroneous traversability prediction can become more apparent, increasing the risk of unrecoverable rover's wheel slip and immobilization. In this work, we propose a new path planning algorithm that explicitly accounts for such erroneous prediction. The key idea is the probabilistic fusion of distinctive ML models for terrain type classification and slip prediction into a single distribution. This gives us a multimodal slip distribution accounting for heterogeneous terrains and further allows statistical risk assessment to be applied to derive risk-aware traversing costs for path planning. Extensive simulation experiments have demonstrated that the proposed method is able to generate more feasible paths on heterogeneous terrains compared to existing methods.

\end{abstract}

\section{Introduction}

Reliable rover autonomy is crucial to exploring celestial surfaces, as manual rover operation from Earth involves significant communication latency. Past rovers, such as those for missions on Mars~\cite{goldberg2002stereo}, were thus equipped with onboard systems that perceive surrounding environments from stereo vision, assess geometric obstacles, and navigate themselves on collision-free trajectories. Nevertheless, many human interventions were required to assess \emph{traversability} in unexplored environments, resulting in slow rover travel. For instance, the Mars Science Laboratory mission reports an average drive distance of the Curiosity rover being limited to 28.9 meters in one Martian solar day (24 hours 39 min~\cite{hecht2009detection}), even though it can travel up to 15.12 m/h~\cite{rankin2020driving}. On extraterrestrial terrain, what turned out to be hazardous for rovers other than apparent obstacles were deformable surfaces. As a known case, the Curiosity rover experienced significant slips on rippled sand at the Hidden Valley, forced to change its route for more solid terrain. Such excessive wheel slips degrade driving speed, increase energy consumption, and eventually cause permanent entrapment in loose, granular materials. Hence, reliable traversability assessment on deformable terrains is essential for autonomous rover operation and, in turn, for faster, more extended rover exploration.

\looseness=-1
Traversability assessment for deformable terrains is challenging due to complex dependencies between physical properties, surface geometry, and rover mobility mechanisms~\cite{wong2008theory}. This problem becomes more complicated when the terrain is \emph{heterogeneous}, \ie, such dependencies even differ from place to place. Thus, instead of theoretical traversability modeling, existing studies attempt to predict traversability via appearance and geometry information using machine learning (ML) algorithms. The appearance information provides cues of surface characteristics, such as material composition, to symbolically categorize heterogeneous terrains via ML classifiers~\cite{rothrock2016spoc,iwashita2020virtual,brooks2012self,otsu2016autonomous}. The geometry information is also given for learning correlations between terrain geometry and slip behavior to predict traversability for unexplored regions~\cite{cunningham2017locally,endo2021terrain,inotsume2021adaptive,angelova2007learning,cunningham2019improving}.

We here argue that for the sake of safe rover navigation, learning-based traversability assessment requires increased attention to potential error inevitably existing in ML predictions. While recent ML models demonstrate significant improvement for robotic perception and control~\cite{kober2013reinforcement,billard2019trends,grigorescu2020survey}, they cannot guarantee perfect predictions.
This common issue of ML can cause unrecoverable rover immobilization. Moreover, highly accurate traversability prediction is expectedly difficult, particularly for distant planets, due to the presence of unobservable terrain conditions and limited observations for unexplored regions.

\begin{figure*}[t]
    \centering
    \includegraphics[width=0.95\hsize]{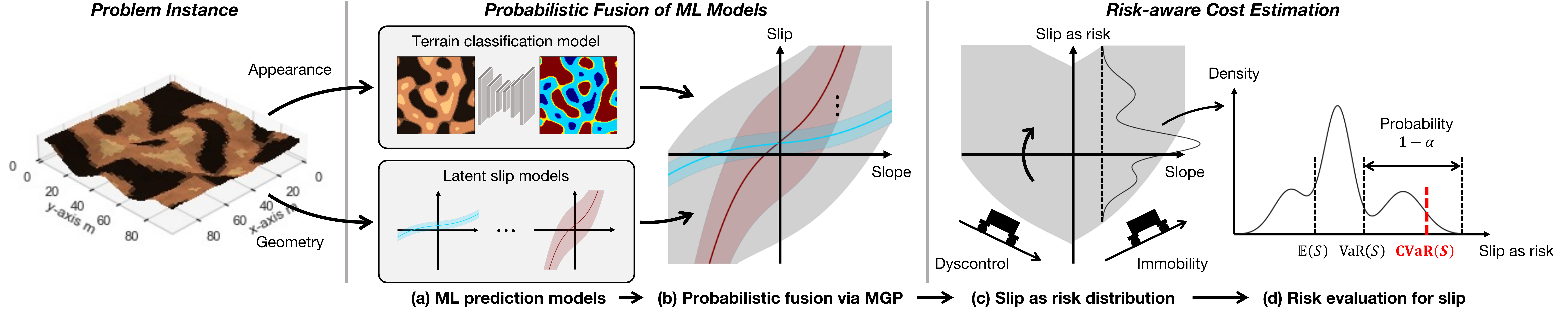}
    \caption{Risk-aware path planning via probabilistic fusion of traversability prediction models and conservative risk evaluation.}
    \label{fig:2}
\end{figure*}

\looseness=-1
To address future \emph{immobilization risk} for safe rover navigation, we devise a path planning algorithm incorporating rational risk assessment against potential error in traversability prediction for heterogeneous terrains. The key idea is the probabilistic fusion of prediction models while taking into account their uncertainties. Suppose we have a terrain classifier  predicting heterogeneous terrain classes given appearance information and latent slip (LS) models mapping geometric features to slip behaviors depending on terrain classes via Gaussian processes (GPs)~\cite{williams2006gaussian} (Fig~\ref{fig:2}a). Our strategy then integrates multiple LS models of all possible terrain classes to form a single, multi-modal slip distribution as mixtures of Gaussian processes (MGP; Fig.~\ref{fig:2}b). This retains the prediction uncertainties of each model and enables us to effectively evaluate the immobilization risk via conditional value at risk (CVaR)~\cite{majumdar2020how} for conservative path planning (Fig.~\ref{fig:2}c-d).

We extensively evaluate our algorithm in synthetic celestial environments, which pose various difficulties in the traversability prediction. Comparison with different prediction models and risk evaluation metrics validates that our algorithm can plan feasible and safer paths on heterogeneous deformable terrains.

\section{Related Work}

Numerous studies have tackled terrain traversability prediction by employing ML algorithms with visual and geometric information.
Classification-based approaches categorically distinguish terrain types, typically traversable or non-traversable regions, from appearance~\cite{rothrock2016spoc} and infrared~\cite{iwashita2020virtual} imagery. Combining vision and proprioceptive (\eg, vibration) sensing has also been studied for increasing robustness against appearance changes~\cite{brooks2012self, otsu2016autonomous}.
Regression-based approaches assess traversability to a more detailed degree.
These approaches model \mbox{(non-)traversability} as the degree of wheel slip and estimate a mapping from terrain inclination to slip via regression algorithms such as GPs~\cite{cunningham2017locally, endo2021terrain, inotsume2021adaptive, angelova2007learning, cunningham2019improving}.
To account for multi-modal slip behaviors with regression-based approaches, classify-then-regress-based methods first classify terrains  from appearance imagery and then exploit class-dependent regressors for geometry-based slip prediction~\cite{cunningham2017locally, endo2021terrain, inotsume2021adaptive}. Other work exploits a mixture-of-experts (MoE) framework to incorporate terrain class likelihoods for multi-modal slip prediction~\cite{angelova2007learning, cunningham2019improving}.

These traversability prediction techniques are further utilized in rover navigation studies for planetary exploration.
For example, Ono~\emph{et al.}~\cite{ono2015risk} use a terrain classifier to detect non-traversable regions for collision-free path planning using the A* algorithm.
Similarly, a path planner by Helmick~\emph{et al.}~\cite{helmick2009terrain} exploits a traversability map predicted by an MoE-based regressor. 
Such direct use of traversability prediction results for rover navigation, however, exposes rovers to the immobilization risk owing to ML-inherent prediction error.

Such prediction error in ML has also been studied under the concept of \emph{uncertainty}.
When quantized via probabilistic models, uncertainty provides useful risk assessment tools in ML applications.
Particularly in the context of risk-aware path planning for planetary environments, planners based on chance constraint formulations have been proposed by exploiting the variance of GPs as traversability prediction uncertainties~\cite{inotsume2020robust,candela2022approach,endo2022active}.
However, these methods rely on single GP models and thus assume homogeneous materials covering given environments or accurate terrain classification results in advance.
Moreover, although chance constraints are suitable for capturing risk corresponding to boolean events (\eg, collisions)~\cite{majumdar2020how}, the immobilization risk incurs a range of costs corresponding to the degree of slip.

Conditional value at risk (CVaR), defined to provide a pessimistic estimate of a random variable by accounting for its uncertainty, is another statistical technique for risk assessment.
CVaR satisfies certain axioms required for rational risk assessment for robotics applications under uncertainty~\cite{majumdar2020how}.
For risk-aware planning, CVaR has been applied to avoid various risks such as moving obstacles in robot navigation~\cite{hakobyan2019risk, ren2023chance}, uncertain regions in semantic segmentation of road scenes~\cite{sharma2020risk}, or an abstract risk in a hazardous process~\cite{barbosa2021risk}. While these studies focus on either collision avoidance~\cite{hakobyan2019risk, ren2023chance}, classification-based traversability modeling~\cite{sharma2020risk}, or a single-modal risk behavior~\cite{barbosa2021risk}, our focus is on handling the immobilization risk that requires continuous and multi-modal modeling of rover slip on various kinds of deformable terrains.
Risk-aware off-road navigation by Cai~\emph{et~al.}~\cite{cai2022risk} uses CVaR to estimate pessimistic robot speed in environments with dirt and vegetation.
Although their use of CVaR for velocity prediction is similar to ours,  their semantic-based scene understanding without considering terrain geometry is unsuitable for capturing the immobilization risk owing to rugged celestial surfaces.

In summary, our work focuses on path planning of rovers in the face of the immobilization risk due to heterogeneous and deformable terrains. 
Compared to existing planners for planetary rover navigation~\cite{ono2015risk,helmick2009terrain,inotsume2020robust,candela2022approach,endo2022active}, our method simultaneously considers multi-modal slip behaviors due to heterogeneous terrains and the immobilization risk. The proposed solution (\ie, MGP-based traversability model integrating visual and geometric information as well as CVaR-based continuous (non-boolean) risk assessment) addresses challenges that are only partly handled by existing CVaR-based risk-aware planners in other areas~\cite{hakobyan2019risk, ren2023chance,sharma2020risk,barbosa2021risk,cai2022risk}.

\section{Preliminaries}

This section describes a general path planning formulation, GPs, and CVaR as technical foundations of our algorithm.

\subsection{Path Planning}

The path planning objective is to find a sequence of feasible robot state transitions that navigates a robot to a destination safely and efficiently. Let $\mathcal{G}=(\mathcal{V},\mathcal{E})$ be a grid map representing an environment in a 2D space, where $\mathcal{V}$ is a finite set of vertices $v$ enumerating possible robot states and $\mathcal{E}$ is a set of edges representing state transitions from each vertex to its eight neighbors. Each edge is associated with a strictly positive cost calculated by a movement cost function $f_{\text{cost}}(v,v')$. The optimal path planning problem is formulated to find a path $\mathcal{P}=\left\{v_1, v_2,..., v_{|\mathcal{P}|}\right\}$, from the start $v_1=v_{\text{start}}\in\mathcal{V}$ to the goal $v_{|\mathcal{P}|}=v_{\text{goal}}\in\mathcal{V}$, with the minimum possible total cost defined as follows:
\begin{equation}
\label{eq:pathplanning}
\min_\mathcal{P} \sum_{i=1}^{|\mathcal{P}|-1}f_{\text{cost}}\left(v_i, v_{i+1}\right).
\end{equation}

\subsection{Gaussian Process for Traversability Modeling}
\label{section:gp}
Wheeled rovers experience slip when traveling on deformable terrains. We consider the longitudinal slip ratio $s$ to quantitatively describe the rover traversability as follows:
\begin{equation}
\label{eq:slip_definition}
{s}=\begin{cases}
    \left({u}_{\mathrm{ref}} - {u}\right)/{u}_{\mathrm{ref}},&{u} \leq {u}_{\mathrm{ref}}: \text{driving},\\
    \left({u}_{\mathrm{ref}} - {u}\right) / {u},&{u} > {u}_{\mathrm{ref}}: \text{braking},
\end{cases}
\end{equation}
where ${u}$ and ${u}_{\mathrm{ref}}$ are the actual and reference velocities in the longitudinal direction, respectively~\cite{wong2008theory}. A positive $s$ represents a traverse slower than commanded, with $s=1$ representing complete rover entrapment in deformable terrains. Conversely, a negative $s$ expresses a traverse faster than commanded.

To model the unknown relationship between terrain geometry and wheel slip $s$, we exploit GP, a non-parametric regression approach employing statistical inference to learn dependencies between points in a dataset~\cite{williams2006gaussian}. 
The use of GPs is justifiable as 1) they can handle the effects of unobservable terrain conditions as uncertainty, and 2) they can express nonlinearity in rover-terrain interactions.
A training dataset for GPs may be collected as rovers measure terrain pitch angles $\phi$ and corresponding longitudinal slips~$s$ in the past traverse experience. 
To account for various slip trends on heterogeneous terrains, suppose terrain classes dominate these slip trends. We then introduce a class-dependent GP that probabilistically models the traversability (\ie, slip $s$) at an edge $e\in\mathcal{E}$ for a terrain class $c$ as
\begin{equation}
\label{eq:gp}
    \mathbb{P}_{c}\left(s|\phi_e\right) := \mathcal{N}\left(\mu_c(\phi_e;\boldsymbol{\phi},\boldsymbol{s}), \sigma_c^2(\phi_e;\boldsymbol{\phi},\boldsymbol{s})\right),
\end{equation}
where the predictive mean $\mu_c$ and and variance $\sigma_c^2$ are parameterized with training samples of input pitch angles $\boldsymbol{\phi}$ and output slip measurements $\boldsymbol{s}$ (see \cite{williams2006gaussian} for details).

\subsection{Conditional Value at Risk}
\label{section:cvar}
\looseness=-1
Rovers must measure uncertainty in traversability to assess risk during planning. Unlike collision risk considered in typical planning problems~\cite{luders2013robust}, our immobilization risk by deformable terrains gradually increases with slip magnitude. We exploit CVaR to quantify such risk through probabilistic traversability modeling. CVaR measures a conservative expected value accounting for tail events in a given distribution, enabling rational risk assessment without being too optimistic or pessimistic~\cite{majumdar2020how}. Let $S$ be a random variable of the longitudinal slip ratio at an arbitrary rover pitch. $\text{CVaR}$ at level $\alpha \in \left[0,1\right]$ denotes the expected value of $S$ in the upper $\left(1-\alpha\right)$-tail, subject to the conditional distribution of $S$ as follows:
\begin{equation}
    \label{eq:cvar_plus}
    \text{CVaR}_{\alpha}(S): = \mathbb{E}\left[S|S > \text{VaR}_{\alpha}(S)\right],
\end{equation}
where $\text{VaR}_{\alpha}$ is the value at risk (VaR) expressing $(1-\alpha)$-quantile of $S$ as $\text{VaR}_{\alpha}(S): = \min\left\{s|\mathbb{P}\left(S>s\right)\leq\alpha\right\}$.
Intuitively, CVaR provides pessimistic risk estimates from the given traversability prediction, as illustrated in Fig.~\ref{fig:2}d, factoring in its uncertainty for risk-aware path planning. Given $\alpha$ controls risk assessment behavior from aggressive to conservative throughout the above formulations. If $\alpha=0$, CVaR returns the expectation. If $\alpha=1$, the CVaR outcome matches the worst case.

\section{Risk-aware Path Planning Algorithm}

\subsection{Overview}

This section presents the unified algorithm translating ML-based traversability predictions and their uncertainties as risk-aware traveling costs evaluating immobilization risk in planning. Unlike typical path planning problems, such as shortest pathfinding in 2D maps, our formulation for rover traveling requires variable traversability costs to account for slip behaviors on heterogeneous deformable terrains. Although the actual slip trend is unknown, it chiefly depends on the material composition and surface geometry in given environments. We thus associate the environment graph with appearance and geometry information, particularly the colors and 3D positions of terrain surfaces, to be utilized for traversability cost prediction via ML models. Projecting such information into the graph vertices is a durable assumption for actual exploration missions, as the HiRISE camera used for Mars exploration can capture overhead imagery with at most 0.25 m/pixel resolution~\cite{mcewen2007mars}.

\looseness=-1
The overall method pipeline is illustrated in Fig.~\ref{fig:2} and summarized as follows.
We employ two kinds of pre-trained models: 1) a terrain classifier to pixel-wisely predict terrain classes from  appearance imagery and 2) GPs to model class-dependent LS functions with geometry information. These models are fused into a single, multi-modal probabilistic slip distribution via mixtures of GPs (MGP)~\cite{tresp2000mixture} (Section~\ref{section:traversabilit_prediction}). We then extract CVaR estimates from this MGP to derive an edge-wise cost for planning (Section~\ref{section:risk_aware_cost_evaluation}). We here essentially translate a probabilistic slip prediction with uncertainty into a deterministic, effectively conservative cost by considering immobilization risk due to the potentially erroneous classifier and GPs. Finally, we perform path planning procedures, such as A*, to find a trajectory that will not induce rover entrapment.

\subsection{Probabilistic Fusion of ML Models}
\label{section:traversabilit_prediction}
Here, we elaborate on MGP representing the probabilistic fusion of 
the appearance-based terrain classifier and geometry-based slip regressors (GPs). 
The MGP estimates $\mathbb{P}_{e:v \rightarrow v'}\left(s\right)$, a probability distribution of the slip~$s$ in the transition~$e$ from a vertex $v$ to its neighbor $v'$. 
This distribution is formulated as the sum of multiple class-dependent  GPs  $\mathbb{P}_e\left(s|c\right)$, weighted by classification likelihoods $\mathbb{P}_v(c)$ as 
\begin{equation}
    \mathbb{P}_{e:v \rightarrow v'}\left(s\right) = \sum_{c \in C} \mathbb{P}_v\left(c\right)\mathbb{P}_e\left(s|c\right),
\label{eq:mgp}
\end{equation}
where $\mathbb{P}_e(s|c)=\mathbb{P}_c(s|\phi_e)$ in eq. (\ref{eq:gp}).
This single MGP distribution in eq.~(\ref{eq:mgp}) can explain various slip trends on heterogeneous terrains and can simultaneously express overall uncertainties from both appearance-based terrain classification and geometry-based slip regression.

For terrain classification, any model taking appearance imagery and predicting a categorical distribution at each pixel can be used. In this work, we leverage the U-Net~\cite{ronneberger2015unet}, one of the popular computer vision models for pixel-wise semantic segmentation tasks. 
Applying this model results in per-pixel class probabilities expressed via the softmax function as $\mathbb{P}_{v}(c)={\exp{\left(a_c(v)\right)}}/{\sum_{c' \in C}\exp{\left(a_{c'}(v)\right)}}$, where $a_c(v)$ is the scalar output of U-Net for a vertex $v$ and a terrain class $c$. Note that
selecting a single GP corresponding to the most likely class $c^\star=\arg\max_c \mathbb{P}_v(c)$ in eq.~(\ref{eq:mgp}), instead of summing multiple GPs, reduces the presented traversability prediction to existing one~\cite{cunningham2017locally}. However, such  a formulation ignores classification-owing uncertainties in the risk assessment described next.

\subsection{Risk-aware Cost Estimation} 
\label{section:risk_aware_cost_evaluation}
To perform path planning using eq.~(\ref{eq:pathplanning}), we derive its  traversability costs $f_\text{cost}$ from the slip distribution given by MGP. We design the cost function as \emph{risk-aware travel time}, which considers both traversability in terms of risk endangering rovers and traverse efficiency. Travel time is a desirable criterion for the cost function since it encompasses efficiency and safety in terms of maintaining a higher traverse rate with lower slip. The time to traverse an edge $e:v\to v'$ is given as $t=||\boldsymbol{x}(v)-\boldsymbol{x}(v')|| / u(e)$, where $\boldsymbol{x}(v)$ denotes the 3D position at a vertice $v$ and $u(e)$ is the rover velocity when traversing an edge. Here, $u(e)$ is derived with slip $s$ by converting eq. (\ref{eq:slip_definition}) as follows:
\begin{equation}
    \label{eq:velocity}
    u(e) = 
    \begin{cases}
    \left(1-s\right)u_{\text{ref}},
    &\phi_e \geq 0: \text{ascending},\\
    u_{\text{ref}}/\left(1 + s\right),
    &\phi_e < 0: \text{descending},
    \\
    \end{cases}
\end{equation}
where $\phi_e$ is the pitch angle at the edge $e$.
In the ascending case in eq.~(\ref{eq:velocity}), a larger slip $s\in [0,1]$  can be interpreted as risk since it slows down rover velocity, increases traverse time, and eventually triggers rover immobilization. In the descent case, a negatively larger slip $s\in(-1,0]$ may lead to faster velocity and shorter travel time. However, this seemingly `preferable' state should be actually regarded as another risk because it induces rover dyscontrol, such as deviation from the planned trajectory and tip-over situation.

To comprehensively evaluate these two types of risks from slip phenomena, we introduce \emph{slip as risk. 
Given a random variable of slip $S(\phi)$ parameterized by a pitch angle~$\phi$, the slip as risk is defined as another random variable as}
\begin{equation}
    S_{\text{risk}}(\phi) =
    \begin{cases}
    S(\phi), 
    &\phi \geq 0: \text{ascending},\\
    2S(0) - S(\phi),
    &\phi < 0: \text{descending}.\\
    \end{cases}
    \label{eq:random_variable}
\end{equation}
\emph{Here, the distribution of $S(\phi)$ follows MGP in eq.~(\ref{eq:mgp}).
Essentially, the slip as risk in eq.~(\ref{eq:random_variable}) evaluates the deviation from the  state of traversing on the flat ground (\ie, $\phi = 0$), by vertically flipping $S(\phi)$ in $\phi < 0$ (Fig.~\ref{fig:2}c), assuming that the flat-ground traversing is most stable and least risky.}
We then extract a pessimistic estimate of $S_{\text{risk}}(\phi_e)$ via CVaR as 
\begin{equation}
    s_{\text{risk}}^{+}(e) = \text{CVaR}_{\alpha}\left(S_{\text{risk}}(\phi_e)\right).
\end{equation}
By substituting $s=s_{\text{risk}}^{+}$ into eq.~(\ref{eq:velocity}), we obtain a rover velocity accounting for the comprehensive risks, denoted as $u_{\text{risk}}(e)$. The cost function is derived as the risk-aware travel time that evaluates time efficiency and rover traverse stability in both ascent and descent directions, as follows:
\begin{equation}
    f_\text{cost}(e:v\to v') = \frac{||\boldsymbol{x}(v) - \boldsymbol{x}(v')||}{u_{\text{risk}}(e)}.
\end{equation}
Note that MGP in eq. (\ref{eq:mgp}) can be alternatively expressed with $\mathbb{P}_{v'}(c)$ instead of $\mathbb{P}_{v}(c)$. We thus compute costs in two ways using both definitions and use their average for planning.

\section{Simulation Experiments}

This section demonstrates the proposed path planning algorithm through extensive simulations and validates its robustness against unreliable traversability prediction. 

\subsection{Datasets}
We prepare three synthetic datasets for path planning problems with varying difficulties in traversability prediction. Each dataset has the training, validation, and testing problem instances as 96 $\times$ 96 grid maps with a resolution of one grid per meter, providing RGB and height maps as inputs for planners. As hidden information, each instance has pixel-wise assignment of terrain classes $c \in C$ and a set of class-dependent slip functions $f_{c}(\phi)$. Throughout the dataset creation, fractal terrain modeling \cite{yokokohji2004evaluation} is applied to generate randomly distributed rough terrain geometry. 

To synthesize diverse but reasonably controlled datasets, we here define \emph{environment groups},  a random subset of terrain classes with occupancy ratios for controlling the terrain class occurrence in each map (\eg, 50\% for class \#1, 30\% for class \#4, and 20\% for class \#6). Training, validation, and testing subsets individually have ten distinct environment groups (\ie, ten different random choices of terrain classes), from each of which we generate 100, 50, and 10 problem instances, respectively. This results in 1000 training, 500 validation, and 100 testing instances. For each problem instance, we first synthesize a random Perlin noise image, cluster the values based on the occupancy ratios specified by the environmental group, and assign a corresponding terrain class (\ie, slip model) to each cluster. 
For each slip model, a slip measurement process is simulated via additive zero-mean Gaussian noise of a pre-defined variance as $s=f_c(\phi) + \epsilon_c$. Synthesized slip measurements are used to train each class-dependent GP. In-situ measurements are also taken during path executions by substituting calculated $\phi_e$ into $s$.

\begin{figure}[t]
    \adjustbox{valign=t}{
    \begin{minipage}[t]{0.25\linewidth}
        \centering
        \includegraphics[clip, height=22mm]{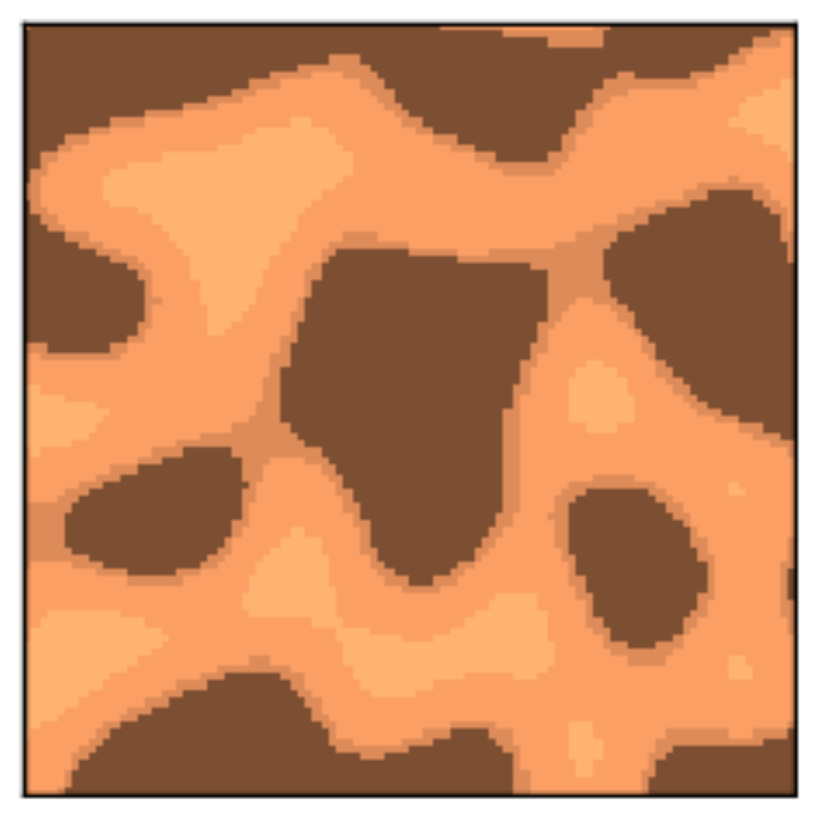}
    \end{minipage}}
    \hspace{-8pt}
    \adjustbox{valign=t}{
    \begin{minipage}[t]{0.25\linewidth}
        \centering
        \includegraphics[clip, height=22mm]{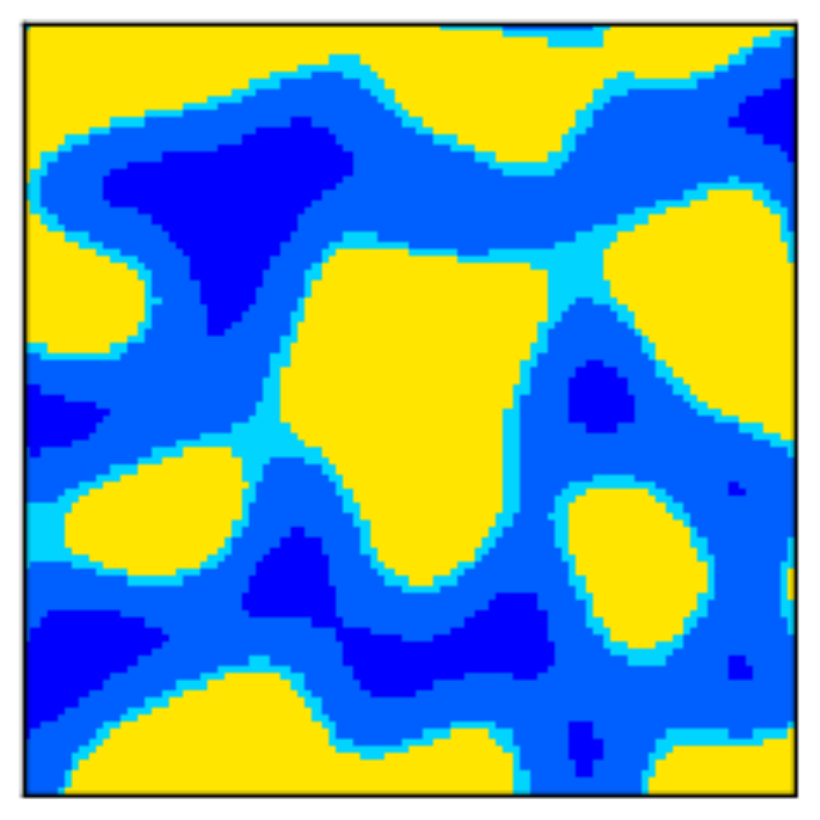}
    \end{minipage}}
    \hspace{-8pt}
    \adjustbox{valign=t}{
    \begin{minipage}[t]{0.40\linewidth}
        \centering
        \includegraphics[clip, height=25mm]{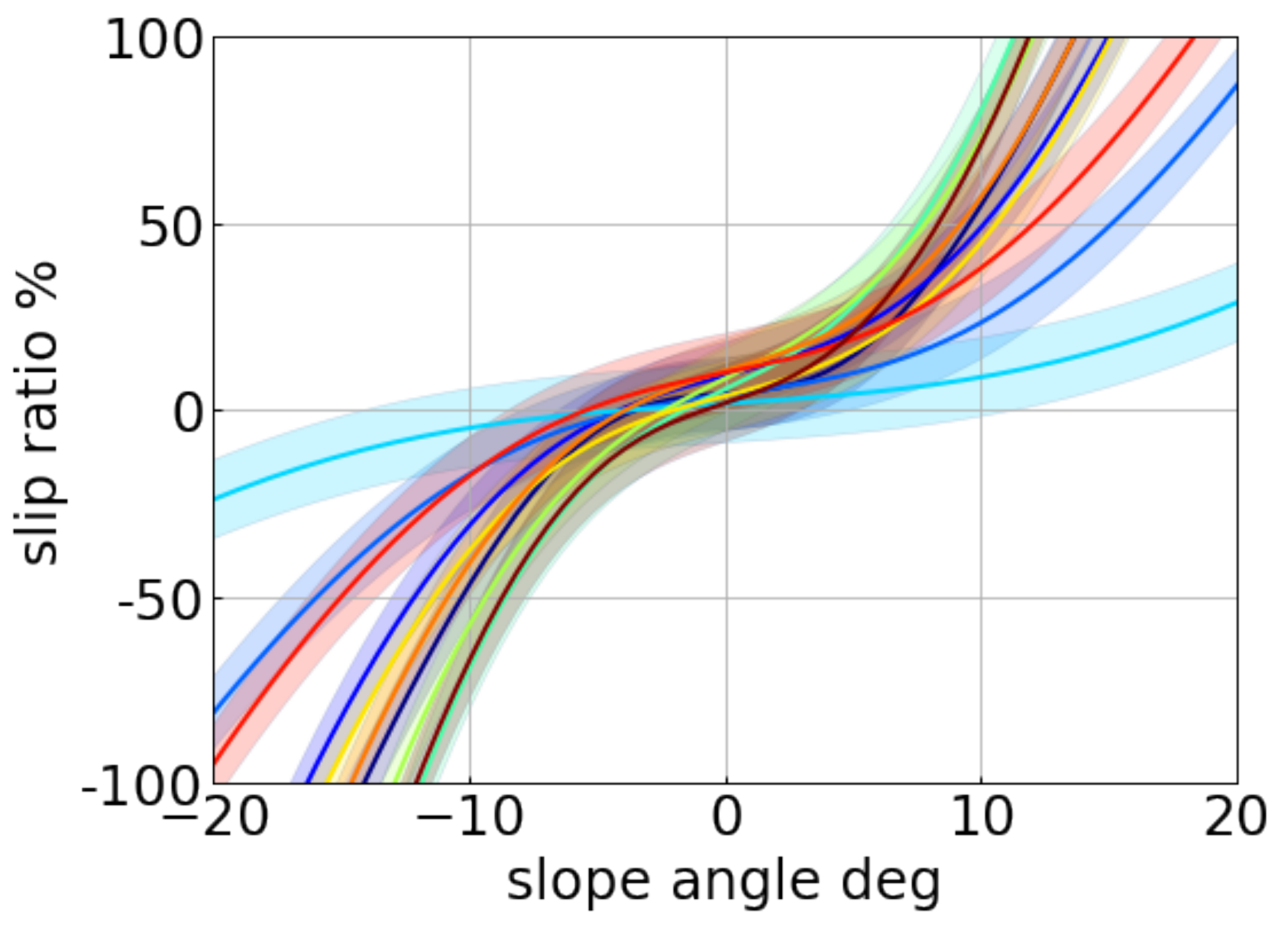}
    \end{minipage}}
    \begin{minipage}[t]{0.28\linewidth}
        \subcaption{}
    \end{minipage}
    \hspace{-8pt}
    \begin{minipage}[t]{0.31\linewidth}
        \subcaption{}
    \end{minipage}
    \hspace{-8pt}
    \begin{minipage}[t]{0.43\linewidth}
        \subcaption{}
    \end{minipage}\\

    \vspace{-6pt}
    
    \adjustbox{valign=t}{
    \begin{minipage}[t]{0.25\linewidth}
        \centering
        \includegraphics[clip, height=22mm]{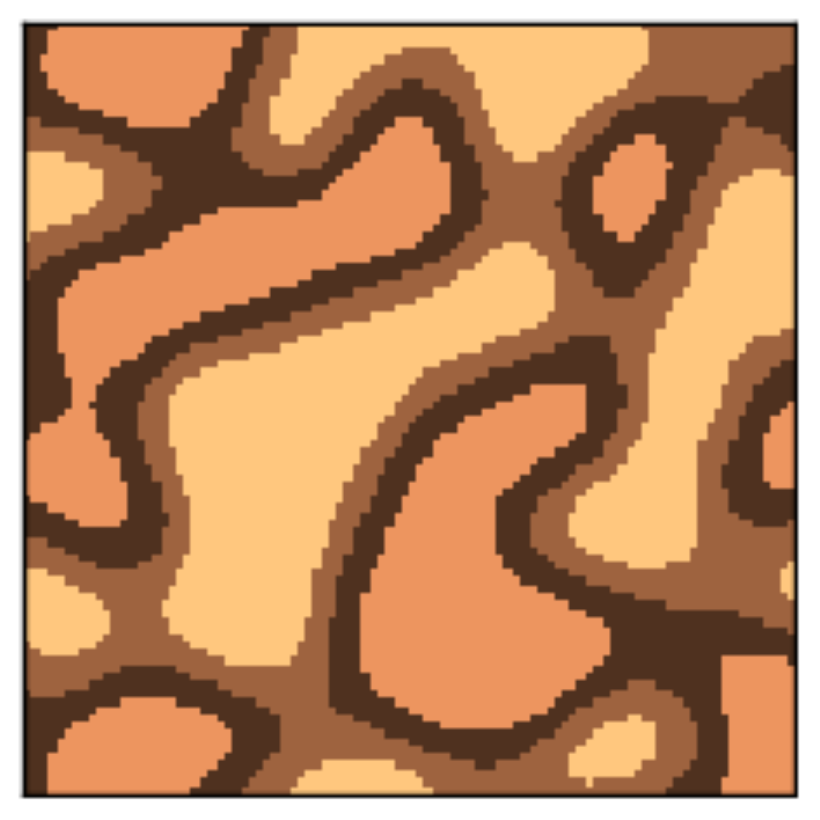}
    \end{minipage}}
    \hspace{-8pt}
    \adjustbox{valign=t}{
    \begin{minipage}[t]{0.25\linewidth}
        \centering
        \includegraphics[clip, height=22mm]{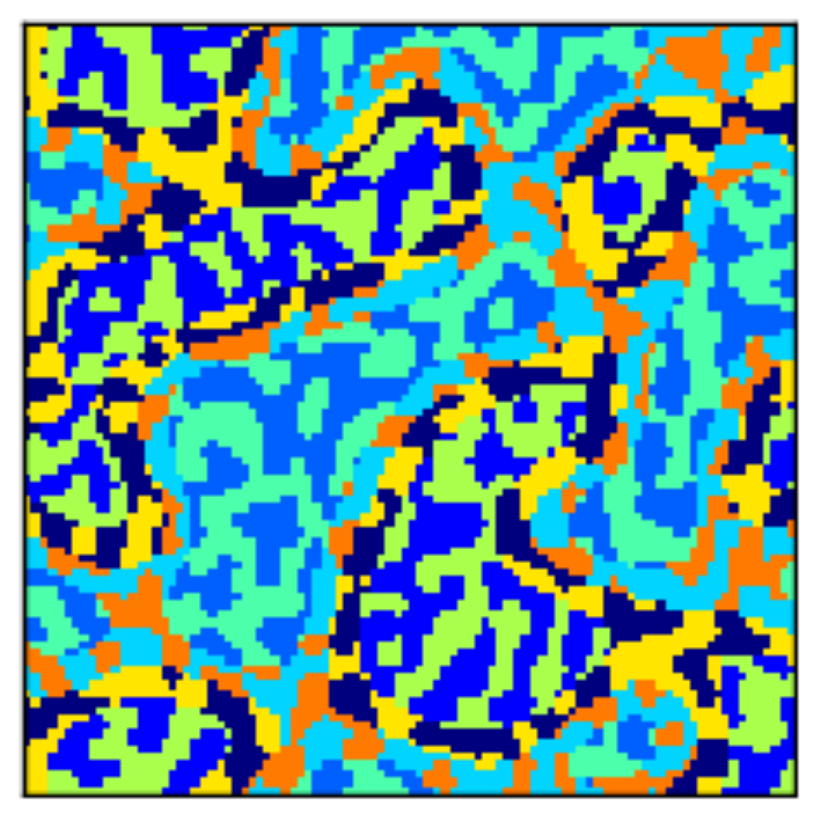}
    \end{minipage}}
    \hspace{-8pt}
    \adjustbox{valign=t}{
    \begin{minipage}[t]{0.40\linewidth}
        \centering
        \includegraphics[clip, height=25mm]{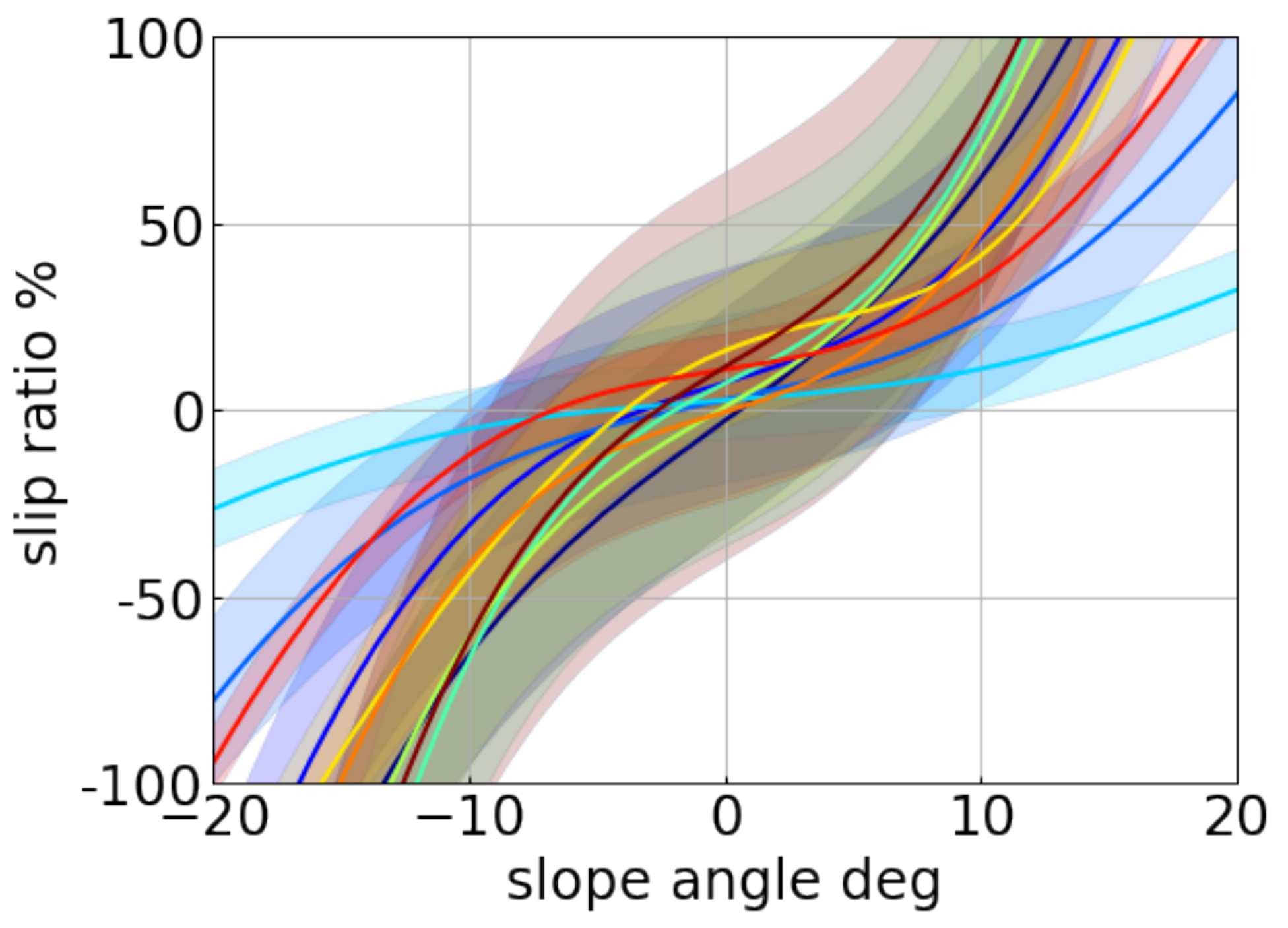}
    \end{minipage}}
    \begin{minipage}[t]{0.28\linewidth}
        \subcaption{}
    \end{minipage}
    \hspace{-8pt}
    \begin{minipage}[t]{0.31\linewidth}
        \subcaption{}
    \end{minipage}
    \hspace{-8pt}
    \begin{minipage}[t]{0.43\linewidth}
        \subcaption{}
    \end{minipage}\\
    \vspace{-10pt}
    \caption{Visualized examples of the (a)-(c) Std dataset and (d)-(f) AA dataset. Each dataset consists of color maps, corresponding terrain class maps, and GPs expressing learned latent slip functions (solid line: mean, shaded area: 95\% confidence region). Color in terrain class maps and GPs denote distinct terrain classes.}
    \label{fig:datasets}
\end{figure}

With the above approach in mind, here is the detail of the three datasets we have created.
\begin{itemize}
\item \textbf{Standard (Std) Dataset:}
This dataset adopts a simple setting where pixel color information identifies terrain classes and slip measurements contain slight noise $\epsilon_c$. Each environment group consists of four out of ten terrain classes. Fig.~\ref{fig:datasets} (a) and (b) show an example terrain appearance image and a corresponding terrain class map coded by distinctive colors, respectively. Fig.~\ref{fig:datasets} (c) shows learned slip models via GPs.
\item \textbf{Erroneous Slip (ES) Dataset:}
A more complicated setting is considered in this dataset by imposing greater noise $\epsilon_c$ in slip measurements as the gradient magnitude of $f_{c}(\phi)$ increases. 
Hence, learned GPs result in erroneous slip models shown in Fig.~\ref{fig:datasets} (f). As described in \cite{cunningham2017locally}, such erroneous slip modeling can happen due to multiple factors, including insufficient slip measurements and unobservable terrain conditions.
\item \textbf{Ambiguous Appearance (AA) Dataset:}
We replicate another complicated setting where multiple terrain classes appear with the same appearance, assuming situations, \eg, when distinctive terrain characteristics do not provide visually identifiable cues or one terrain material is covered by another. This dataset comprises four pairs of terrain classes, each with high-variance GPs (as in the ES dataset) but sharing the same color cues, resulting in eight classes with less appearance diversity. Each environment group is then defined by the random ordering of those classes such that four different colors always appear, as shown in Fig.~\ref{fig:datasets} (d) and (e).
\end{itemize}

\subsection{Implementation Details}

As the path search algorithm, we use the A* search, which returns a resolution-optimal solution as long as it exists in a given discrete domain. As the terrain classifier, we adopt the U-Net model~\cite{ronneberger2015unet} with the ResNet-18 encoder~\cite{he2016deep} to provide classification likelihoods. This model was trained for each dataset separately using its training subset for 50 epochs using the Adam optimizer~\cite{kingma2015adam} with a learning rate of 0.001 and a batch size of 8. The model weight producing the lowest loss on the validation subset was used for evaluation on the test subset. Computing VaR and CVaR for random variables associated with the MGP distribution in eq.~(\ref{eq:mgp}) was carried out approximately through a Monte Carlo simulation.

\begin{figure*}[t!]
    \centering
    \includegraphics[width=\hsize]{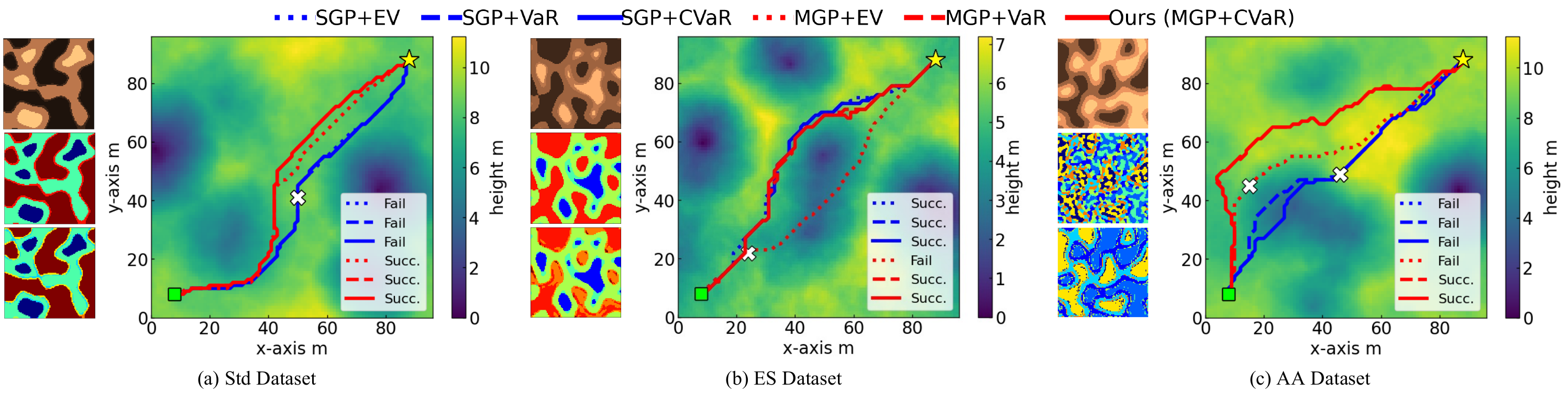}
    \vskip -2mm
    \caption{Path planning results over different datasets. In the environments represented as heatmaps, green squares, and yellow stars are start and goal positions. White crosses denote positions where rovers fail to transit to the next states due to experiencing $s=1$. Input colors, ground truth terrain classes, and terrain classification results are shown from top to bottom on the left side of each heatmap.}
    \label{fig:4}
\end{figure*}

\subsection{Experimental Setups}

In every problem instance, planners search the path from the start position (8~m, 8~m) to the goal (88~m, 88~m). After finding a solution, a rover follows the given path at $u_{\text{ref}} =$ 0.1 m/sec while receiving an edge-wise noisy slip $s_e$. We regard the solution as a success path if we never observe $s_e \geq$ 1 (complete rover immobilization) and $s_e \leq-1$ (uncontrollable situations such as tip-over) on the way. The following four metrics quantify algorithm performances.
\begin{itemize}
    \item \textbf{Solved rate (Sol.) [\%]} expresses the ratio of problem instances for which a planner finds solutions. 
    \item \textbf{Success rate (Succ.) [\%]} denotes the ratio of problem instances whose found solutions are successful.
    \item \textbf{Total time ($T_{\text{total}}$) [min]} provides path efficiency by calculating the total traversing time along the path when traversing is successful.
    \item \textbf{Maximum slip ($s_{\text{max}}$) [\%]} evaluates path safety against rover entrapment by calculating $100 \times \max(\boldsymbol{s}_e)$ after path execution, where $\boldsymbol{s}_e$ denotes the experienced slips. A lower $s_{\text{max}}$ indicates a safer traverse on terrains.
\end{itemize}
To see the effectiveness of our proposed approach, we also evaluated different traversability prediction models and risk evaluation metrics. As for the prediction, a single Gaussian process (SGP) approach is given by selecting a GP based on classified terrain types~\cite{cunningham2017locally}. In addition to VaR and CVaR, the expected value (EV) quantifies the given slip distribution for cost calculation. With combinations of the above metrics, planning procedures run on the datasets for comparison.

\begin{table}[t]
    \centering
    \caption{Quantitative path planning results over different datasets}
    \scalebox{0.95}{
    \begin{tabular}{rlcccc}
        \toprule
        \multicolumn{6}{c}{\textbf{Standard (Std) Dataset}}\\
        \hline
         && Sol. & Succ. & $T_{\text{total}}$ & $s_{\text{max}}$ \\
        \hline
        SGP~\cite{cunningham2017locally}&+EV & 100 & 71 & 22.6 $\pm$ 3.7 & 58.3 $\pm$ 47.0 \\
        &+VaR & 100 & 74 & 22.6 $\pm$ 3.9 & 55.6 $\pm$ 45.4 \\
        &+CVaR & 100 & 75 & 22.6 $\pm$ 3.9 & 55.3 $\pm$ 45.8 \\
        \hline
        MGP&+EV & 100 & 75 & \textbf{22.5} $\pm$ 3.5 & 56.1 $\pm$ 44.4 \\
        &+VaR & 100 & 93 & 22.9 $\pm$ 4.4 & 47.5 $\pm$ 30.7 \\
        \hline
        \multicolumn{2}{c}{\textbf{Ours (MGP+CVaR)}} & 100 & \textbf{96} & 23.2 $\pm$ 5.2 & \textbf{45.9} $\pm$ 24.2 \\
        \bottomrule
        \toprule
        \multicolumn{6}{c}{\textbf{Erroneous Slip (ES) Dataset}}\\
        \hline
         && Sol. & Succ. & $T_{\text{total}}$ & $s_{\text{max}}$ \\
        \hline
        SGP~\cite{cunningham2017locally}&+EV & 100 & 60 & \textbf{23.0} $\pm$ 3.5 & 70.3 $\pm$ 40.7 \\
        &+VaR & 98 & 65 & 24.3 $\pm$ 5.1 & 68.8 $\pm$ 44.7 \\
        &+CVaR & 93 & 62 & 25.0 $\pm$ 5.7 & 67.9 $\pm$ 44.8 \\
        \hline
        MGP &+EV & 100 & 71 & 23.1 $\pm$ 4.4 & 68.2 $\pm$ 38.2 \\
        &+VaR & 98 & \textbf{77} &25.2 $\pm$ 12.7 & 63.1 $\pm$ 38.8 \\
        \hline
        \multicolumn{2}{c}{\textbf{Ours (MGP+CVaR)}} & 90 & \textbf{77} & 27.2 $\pm$ 37.4 & \textbf{60.5} $\pm$ 37.1 \\
        \bottomrule
        \toprule
        \multicolumn{6}{c}{\textbf{Ambiguous Appearance (AA) Dataset}}\\
        \hline
         && Sol. & Succ. & $T_{\text{total}}$ & $s_{\text{max}}$ \\
        \hline
        SGP~\cite{cunningham2017locally} &+EV & 100 & 11 & \textbf{22.1} $\pm$ 1.7 & 92.9 $\pm$ 27.4 \\
        &+VaR & 100 & 7 & 24.8 $\pm$ 4.0 & 95.8 $\pm$ 19.3 \\
        &+CVaR & 100 & 2 & 26.7 $\pm$ 1.8 & 96.3 $\pm$ 19.3 \\
        \hline
        MGP &+EV & 100 & 38 & 24.0 $\pm$ 5.6 & 81.4 $\pm$ 32.9 \\
        &+VaR & 98 & \textbf{95} & 25.7 $\pm$ 5.3 & 64.5 $\pm$ 22.3 \\
        \hline
        \multicolumn{2}{c}{\textbf{Ours (MGP+CVaR)}} & 96 & \textbf{95} & 26.4 $\pm$ 6.3 & \textbf{63.7} $\pm$ 21.6 \\
        \bottomrule
    \end{tabular}
    }
    \label{tab:quantitative_results}
\end{table}

\subsection{Results}

\subsubsection{Quantitative Comparison}

Table \ref{tab:quantitative_results} summarizes quantitative path planning performances on the three datasets, with $\alpha$ = 0.99 in VaR and CVaR. Throughout the datasets, our proposed approach owns the highest performance in terms of safety to avert immobilized situations, as shown by its success rates and maximum slips during path execution. Having lower slip also contributes to retaining efficiency in time by maintaining rover velocity, as shown by the total time outcomes. These results are preferable for planetary exploration, where safety is of utmost importance, as rover entrapment leads to mission failure while overly conservative rover navigation delays mission schedules. Erroneous slip modeling, despite higher appearance diversity experienced in the ES Dataset, poses a relative performance decrease in planning outcomes compared to the other datasets. We found that the planner with VaR- and CVaR-based risk evaluations sometimes does not provide solutions. This can happen when the goal position is in hazardous areas. Dismissing such risk to prioritize finding solutions worsens the success rate, as indicated by the EV-based planners. The MGP formulation for traversability prediction greatly contributes to the success of rover navigation. The most effective cases are those provided by the AA Dataset, where terrain appearance cannot assure its terrain class. Without MGP, traversability prediction only refers to representative slip models according to identified terrain properties, resulting in failed traverses due to huge prediction errors from wrong slip models. Our proposed algorithm solves this issue by integrating multiple GPs with terrain likelihoods, thus enabling careful risk evaluation accounting for less-likely tail events that can lead to immobilized situations.
Total computation time for planning, including cost estimation with MGP, is typically 10 to 15 minutes on a standard computer in Python implementation.

\begin{figure}[t!]
    \centering
    \includegraphics[width=0.8\hsize]{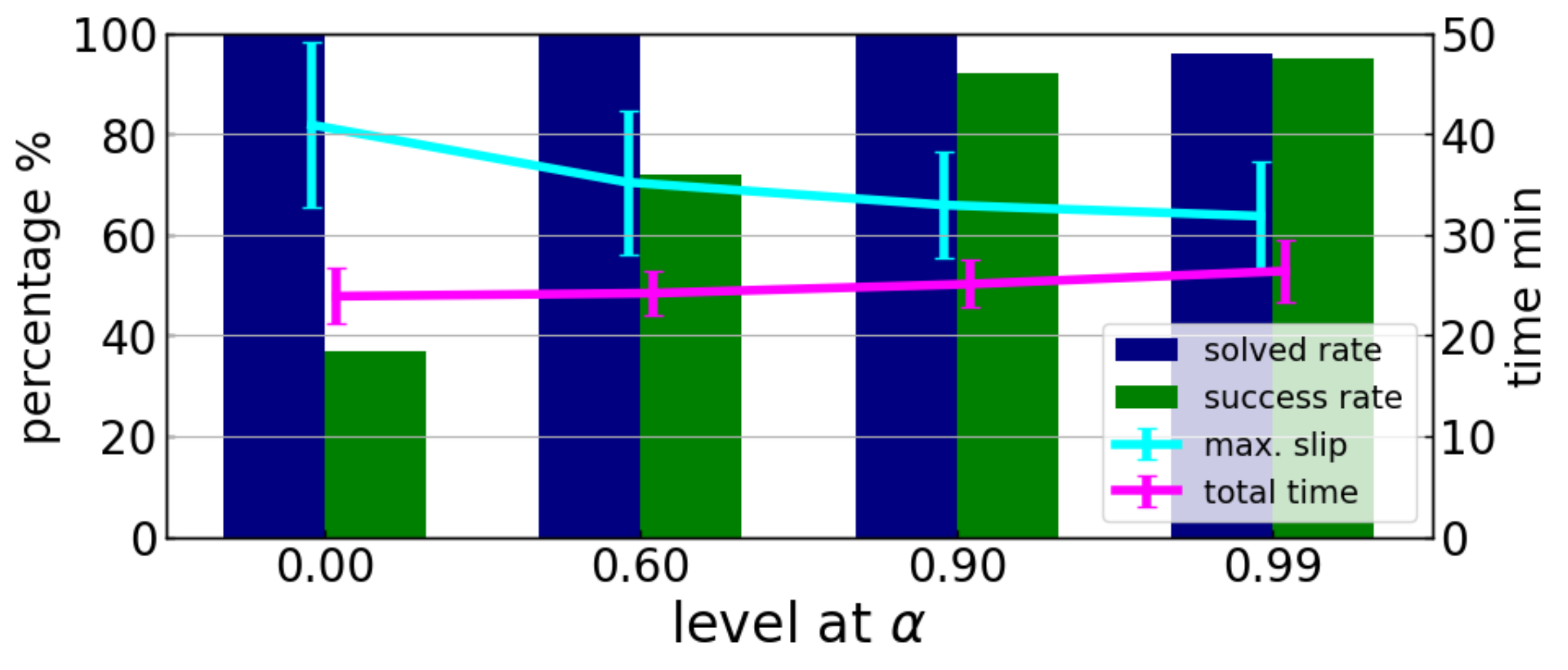}
    \vskip -2mm
    \caption{Parameter study results over AA Dataset. Line graphs use mean and standard deviation format.}
    \label{fig:5}
\end{figure}

\subsubsection{Qualitative Comparison}

Fig. \ref{fig:4} illustrates how the planners with different risk metrics plan paths for a problem instance from each Dataset. MGP-based approaches tend to find trajectories along geometrically benign regions compared to SGP-based ones, but the MGP+EV-based planner often seeks a too optimistic path leading to rover immobilization, as shown in Fig. \ref{fig:4} (b). This is because the MGP generates a slip distribution covering a broader range of slips compared to a single GP. VaR and CVaR are thereby necessary for rational risk evaluation to see less likely occurring slips in the MGP distribution, enabling successful rover traverses in challenging environments.

\subsubsection{Parameter Study}

Finally, a parameter study testing $\alpha  \in \{0.00, 0.60, 0.90, 0.99\}$ for the AA Dataset is conducted to see its effects against the performance of the proposed algorithm. Fig.~\ref{fig:5} indicates that $\alpha$ governs the trade-off between safety versus efficiency: a higher $\alpha$ leads to an increased success rate by navigating rovers on a less-slippery path, while a lower $\alpha$ allows a more aggressive maneuver to shorten the time to traverse. The controllability of planning behavior via simple parameter tuning is a beneficial aspect of our method, considering that priorities of safety and efficiency will change depending on the mission timeline; we may want to prioritize safety during the early exploration phase when most terrain properties are unknown.


\subsubsection{Open Issues and Possible Extensions}

Although our proposed algorithm can significantly improve safely traversing celestial environments, no theoretical guarantee is proven for avoiding rover entrapment. Possible extensions will complementarily exploit probabilistic constraints, such as chance constraints, by formulating reasonable interpretations for non-boolean risk of immobility.

\section{Conclusion}
We presented a new path planning algorithm that can explicitly consider the uncertainty in ML-based traversability prediction for reliable rover operations on heterogeneous deformable terrains. Through the extensive simulation experiments, we confirmed that the proposed algorithm consistently outperformed existing methods in terms of planning success rates and the maximum degrees of wheel slip experienced during path execution. Future work will investigate the feasibility of the proposed method using actual data such as those for Mars~\cite{ono2016data}. 
Another possible future direction is to extend the algorithm to refine the prediction model through online learning~\cite{endo2022active}.

\balance
\bibliographystyle{IEEEtran}
\bibliography{IEEEstyle}

\end{document}